\documentclass[letterpaper, 10 pt, conference]{ieeeconf}  % Comment this line out if you need a4paper

\IEEEoverridecommandlockouts                              % This command is only needed if 
\usepackage{graphics} % for pdf, bitmapped graphics files
\usepackage{epsfig} % for postscript graphics files
\usepackage{xcolor}
\usepackage{amssymb}
\usepackage{adjustbox}
\usepackage{booktabs} % To thicken table lines
\usepackage{amsmath}
\usepackage[ruled,vlined]{algorithm2e}
\usepackage{hyperref}
\newcommand{\CP}[1]{\ignorespaces}
\newcommand{\ie}{\textit{i.e.}}
\newcommand{\eg}{\textit{e.g.}}
\title{\LARGE \bf
AttDLNet: Attention-based Deep Network for 3D LiDAR Place Recognition}

\author{Tiago Barros, Luís Garrote , Ricardo Pereira, Cristiano Premebida, Urbano J. Nunes % 
\thanks{The authors are with the University of Coimbra, Institute of Systems and Robotics, Department of Electrical and Computer Engineering, Portugal.
E-mails:{\tt\small\{tiagobarros,~garrote,~ricardo.pereira,
~cpremebida,~urbano\}@isr.uc.pt}}%
}

\begin{document}

\maketitle
\thispagestyle{empty}
\pagestyle{empty}

%%%%%%%%%%%%%%%%%%%%%%%%%%%%%%%%%%%%%%%%%%%%%%%%%%%%%%%%%%%%%%%%%%%%%%%%%%%%%%%%
\begin{abstract}
%\CP{Place recognition is a fundamental task for SLAM and global localization, which have been using 3D LiDAR data as a key modality for robust operation.} 
LiDAR-based place recognition is one of the key components of SLAM and global localization in autonomous vehicles and robotics applications. With the success of DL approaches in learning useful information from 3D LiDARs, place recognition has also benefited from this modality, which has led to higher re-localization and loop-closure detection performance, particularly, in environments with significant changing conditions. 
Despite the progress in this field, the extraction of proper and efficient descriptors from 3D LiDAR data that are invariant to changing conditions and orientation is still an unsolved challenge. 
To address this problem, this work proposes a novel 3D LiDAR-based deep learning network (named AttDLNet) that uses a range-based  proxy representation for point clouds and an attention network with stacked attention layers to selectively focus on long-range context and inter-feature relationships. The proposed network is trained and validated on the KITTI dataset and an ablation study is presented to assess the novel attention network. Results show that adding attention to the network improves performance, leading to efficient loop closures, and outperforming an established 3D LiDAR-based place recognition approach. From the ablation study, results indicate that the middle encoder layers have the highest mean performance, while deeper layers are more robust to orientation change. The code is publicly available at:\url{ https://github.com/Cybonic/AttDLNet} 

\end{abstract}

%%%%%%%%%%%%%%%%%%%%%%%%%%%%%%%%%%%%%%%%%%%%%%%%%%%%%%%%%%%%%%%%%%%%%%%%%%%%%%%%

\section{INTRODUCTION}

Place recognition has been the focus of much research over the last decade with particular interest by the autonomous vehicle community, which uses place recognition to achieve long-term localization.
Place recognition is a perception-based approach that recognizes previously visited places using visual, structural, and/or semantic cues.  
Although multiple approaches \cite{sunderhauf2015performance,naseer2017semantics} have been proposed for place recognition with promising results in appearance changing and extreme viewpoint variation scenarios, some fundamental problems are still open for research: perceptual aliasing (\ie, places with similar appearance generated from two distinct locations); observations taken over time in the same location that exhibit significant appearance changes due to day-night variation weather, seasonal or structural changes; and viewpoint invariance. 

\begin{figure}[t]
\includegraphics[width=1\columnwidth]{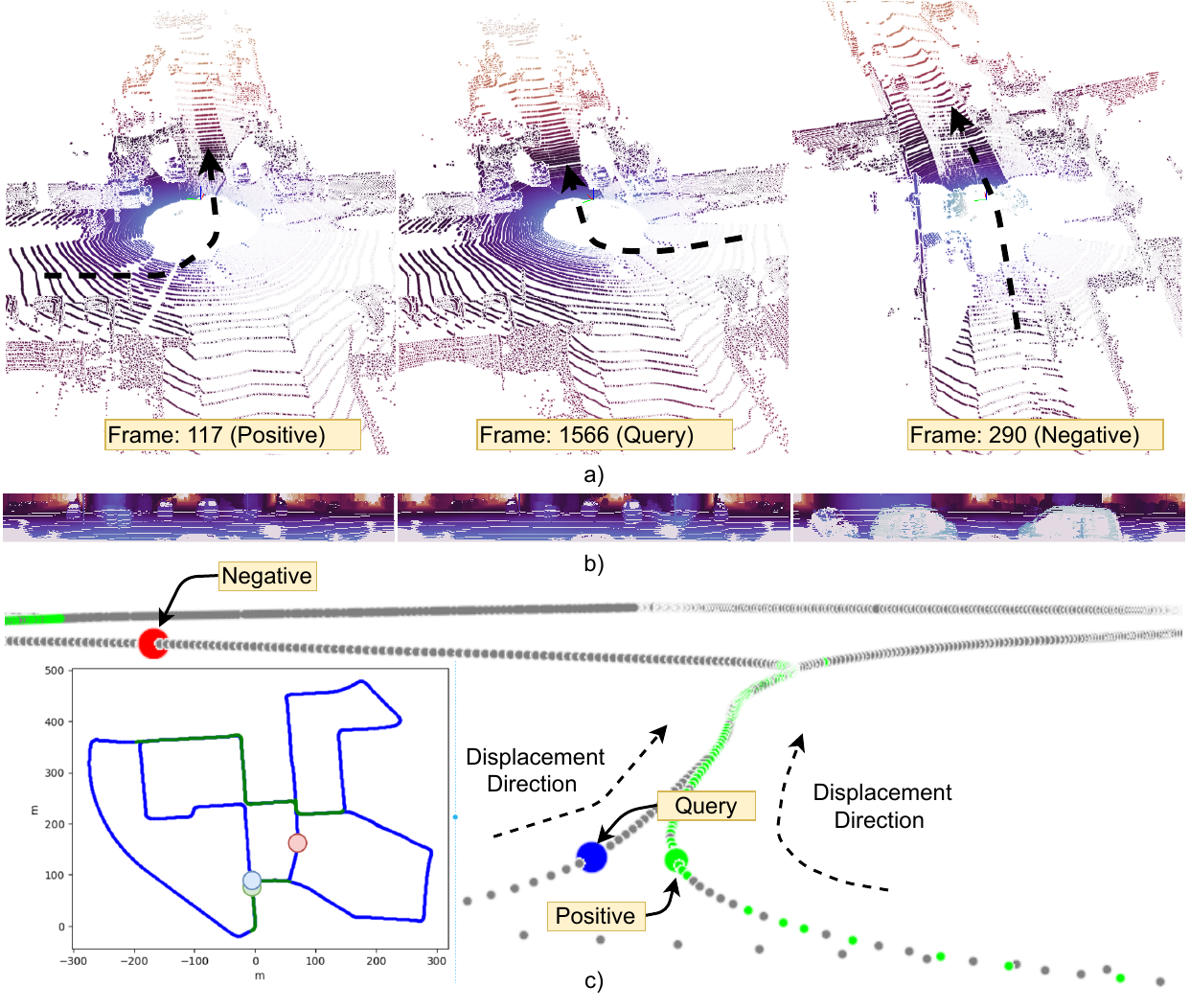}
\caption{Illustration of a query (denoted by the blue-circle), a positive (green-circle) and a negative frame (red-circle) of sequence $00$ - extracted from the KITTI dataset. The above representations correspond to: a) point clouds, b) spherical-based range (proxy) model input, c) geometrical localization.}
\label{fig:triplet}
\end{figure}

The structural information captured by 3D LiDAR sensors has proven to be descriptive for place recognition (as illustrated in Fig. \ref{fig:triplet}) and additionally to be robust to appearance change, which is an advantage compared to vision approaches. A very common approach of generating such global descriptors from point clouds is to use methods such as VLAD (Vector of Locally Aggregated Descriptors), which aggregate local features into global descriptors \cite{angelina2018pointnetvlad}. Another research direction is to formulate the place recognition problem as a graph matching problem, using for example point cloud segmentation approaches \cite{9341060}. However, working with point clouds directly is still computational demanding. Additionally, these graph-based approaches consider each feature equally, disregarding relationships between features or contextual information. A solution to these problems, cf. the approach presented in this work, is to use, on one hand, a proxy representation to alleviate the computational burden and, on the other, use attention mechanisms to exploit feature relationships.

Thus, this work proposes, a novel point cloud-based place recognition approach based on an attention network. The proposed \textbf{Att}ention-based \textbf{D}eep \textbf{L}earning \textbf{Net}work, hereafter called AttDLNet, converts point clouds to a range-based proxy representation (such representation is illustrated in Figure \ref{fig:triplet}) and takes advantage of an attention network with stacked attention layers to learn long-range context and inter-feature relationships.

Succinctly, this work's contribution is an efficient LiDAR-based place recognition approach using attention as a key component to extract rich, efficient, and robust descriptors for real-world environments.   

\begin{figure*}[t]
\centering
\includegraphics[width=1\textwidth, trim={2.3cm 0cm 0cm 0cm},clip]{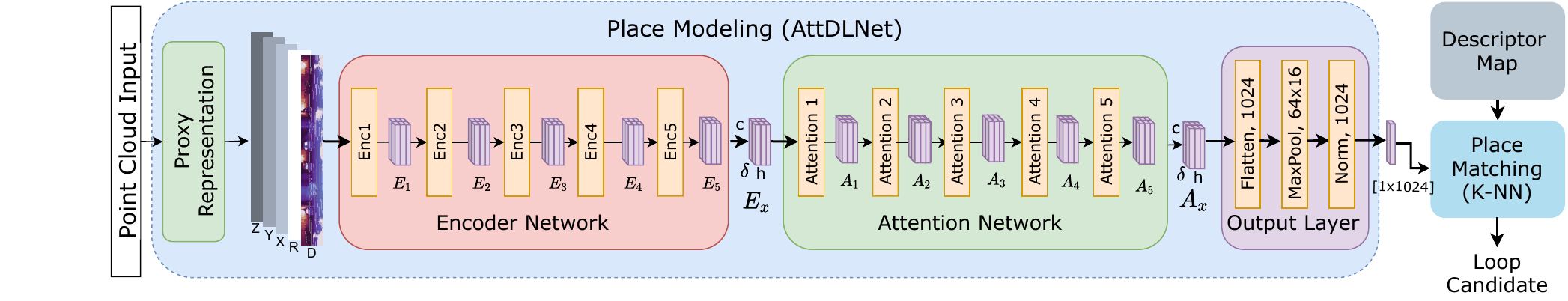}
\caption{Proposed place recognition pipeline with three key modules: place modeling, place matching, and descriptor map. The map contains all descriptors, which are loaded offline. Place matching is computed using a K-nearest neighbor approach that employs cosine similarity for descriptor comparison. Place modeling is computed using AttDLNet: an attention-based network for 3D LiDAR data. AttDLNet receives a point cloud as input; converts the point cloud to a  proxy representation, which is a spherical-based range representation; extracts features from the proxy representation using the encoder network; then calculates relationships among features using an attention network; and finally a descriptor is outputted by output layer, which employs max-pooling and flatten}
\label{fig:overview_pipeline}
\end{figure*}

%%%========= New Section =========
\section{RELATED WORK}

Over the last few years, place recognition has been the subject of much research using DL approaches. Place recognition benefited immensely from the recent developments both in supervised and unsupervised learning, which have been widely applied to vision \cite{wu2019deep}, structural \cite{8968094} or both modalities combined \cite{oertel2020augmenting}. The structural modality, in particular, has gained more attention with the recent DL architectures, which learn features in an end-to-end manner from point clouds.

In point cloud-based DL approaches, the networks can be split into two main categories based on the input representation:  networks that learn directly from point clouds \cite{angelina2018pointnetvlad,9107146}, or networks that learn indirectly from point clouds, learning from proxy representations instead.  Proxy representations are mainly structured-grid representations such as voxels \cite{chang2020spoxelnet}, depth range images \cite{8968094,chen2021overlapnet, chen2020overlapnet}, which are naturally handled by 2D CNNs \cite{8968094} or 3D CNNs \cite{segmap2018} for feature learning. Point clouds, on the other hand, are irregular and unstructured, which makes feature learning, using traditional DL approaches, challenging. In place recognition, place descriptors are learned directly from point clouds using PointNetVLAD \cite{angelina2018pointnetvlad}. PointNetVLAD employs a symmetric max pooling function to aggregate local point features to make the output permutation invariant, which is crucial for place description. Another approach to map raw point clouds directly to descriptors is proposed in \cite{liu2019lpd}, where a graph-based neighborhood aggregation approach is used to extract local structures and reveal the spatial distribution of local features. A graph-based approach is also proposed in \cite{9341060}, which models place recognition as a graph matching problem, using as input segmented point clouds.

More recently, various segmentation, retrieval, and place recognition works have been exploiting spatial and contextual relationships from features using attention mechanisms. In \cite{feng2020point}, a Local Attention-Edge Convolution layer is proposed for the task of segmentation, which leverages local graphs of neighborhood points searched in multi-directions.  Attention is used locally on the graph edges and used globally to learn long-range spatial contextual features. A graph approach is also proposed in \cite{wang2019graph} for the same task, proposing a graph attention convolution where the kernels assume dynamic shapes to adapt to an object's structure. Regarding place recognition, attention is used by PCAN \cite{zhang2019pcan} during the feature aggregation process of local features, which are extracted by an approach inspired on PointNet \cite{qi2017pointnet}. The attention is employed in the NetVLAD \cite{arandjelovic2016netvlad} layer as a context-aware reweighting network to learn multi-scale textual information. A similar approach is proposed in SOE-Net \cite{xia2020soe}, but the local features are modeled based on a point orientation encoder. Contrary to the single modality works aforementioned, PIC-Net \cite{lu2020pic}  uses attention to combine image and point cloud features.

%%%========= New Section =========
\section{PROPOSED APPROACH} \label{sec:pa}

The place recognition approach presented in this work is formulated as a retrieval task, as illustrated in Fig. \ref{fig:overview_pipeline}.  The pipeline comprises three basic modules: place modeling, place matching, and a descriptor map.  The descriptor map is dedicated to maintaining all descriptors, while the place matching module is responsible for returning loop candidates using a k-nearest neighbor (K-NN) approach, which employs a cosine similarity distance for descriptor comparison. These descriptors are generated in the place modeling module, which maps point clouds to a descriptor space, using AttDLNet.

AttDLNet is a deep-learning network that performs the following: converts point clouds to a range-based proxy representation; extracts features from the input using an encoder network; computes feature relationships using an attention network, and converts the features maps from the attention network to adequate descriptors in the output layer.

\subsection{Proxy Representation}

The proxy representation used in this work is a spherical-based range representation inspired by the work in \cite{milioto2019rangenet++}. A point  $p_i = (x,y,z)$, belonging to a point cloud $P$ with a range value of $r$, is projected to spherical coordinate system and then to a image space $(u,v)$ given by:
 \begin{equation}
     {\renewcommand{\arraystretch}{1.5}% for the vertical padding
    \left(
        \begin{array}{l}
          u\\
          v\\
        \end{array}
     \right)
      =
      \left(
        \begin{array}{l}
          \frac{1}{2}[1-\arctan(x,y)\pi^{-1}]\,\omega \\
          \left[1-(\arcsin(z\,r^{-1})+f_{up})f^{-1}\right]\,h \\
        \end{array}
     \right)
     }
    \label{eq:projection}
\end{equation}

\noindent where ($u,v$) corresponds to the pixel coordinates, ($w,h$) are the width and height of the output image tensor,  $f = f_{up} + f_{down}$ and $f_{up}$ are the sensor's total and upper vertical field-of-view, respectively.
 
Thus, a point cloud is projected to a image-based representation with a vertical resolution $h$,  rotational resolution $\omega$, and 5 channels:  one channel for the range values (D), three channels for the $(x, y, z)$ coordinates,  and one channel for remission measurements (R) (\ie, measurements of diffuse reflection). The final tensor with shape $[5 \times h \times \omega ]$ is fed to the encoder network for feature learning.

\subsection{Encoder Network} \label{sec:backbone}
% https://ieeexplore.ieee.org/document/9341060 
The encoder network has the aim of learning features from a range-based proxy representation. The network used in this work is an adapted version of  DarkNet53 \cite{farhadi2018yolov3}, which was originally proposed for image-based object detection \cite{farhadi2018yolov3} and more recently adapted for point cloud-based segmentation in \cite{9341060}, demonstrating in both tasks high levels of descriptiveness. The network comprises several stacked encoder layers (\eg, 2D Convolution, BatchNormalization, and LeakyReLU) with skip connections, receiving as input a tensor with shape [5 $\times$ $h$ $\times$ $\omega$]. The encoder network proposed in \cite{9341060}. which is adopted for this work, has five encoder layers (has illustrated in Fig.\ref{fig:overview_pipeline}); each encoder layer downsamples the previous layer's tensor in the horizontal (rotational) direction, while the vertical shape is kept. The network's performance is studied in Section \ref{sec:ablation}, where an ablation study is conducted to assess the best architecture for place recognition. The network configuration that returns the best performance will be used as the final encoder network, which may differ from the architecture presented in Fig.\ref{fig:overview_pipeline}.

\subsection{Attention Network}
% https://arxiv.org/pdf/1706.03762.pdf 
% https://arxiv.org/pdf/1805.08318.pdf
% Code: https://github.com/heykeetae/Self-Attention-GAN/blob/master/sagan_models.py

% ADD only for layers were choson because of memory constrains 
\begin{figure}[t]
\includegraphics[width=1\columnwidth, trim={0.0cm 0cm 7cm 0cm},clip]{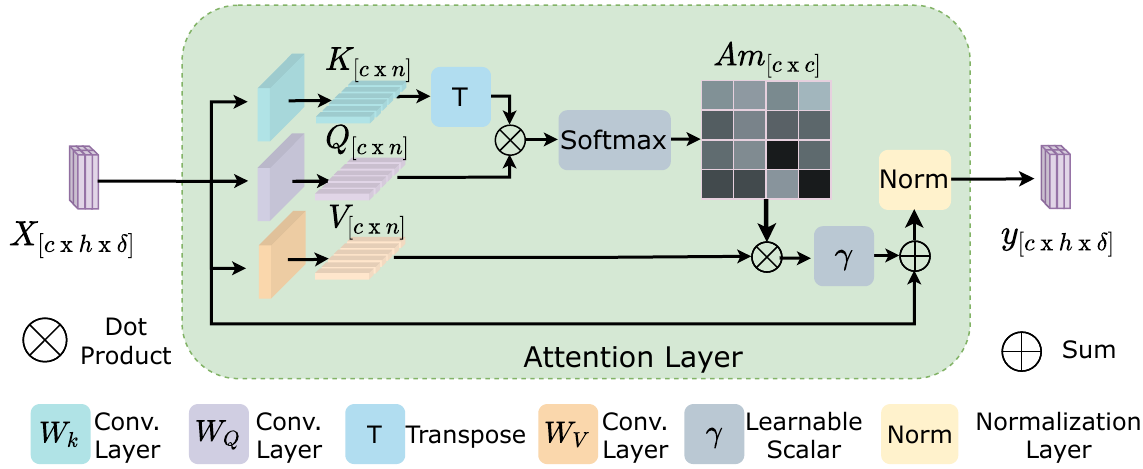}
\caption{Graphical illustration of an attention layer, where \textbf{X} and \textbf{Y} represent, respectively the input and output tensors of the layer. Both tensors have the same dimension.}
\label{fig:attention_network}
\end{figure}

Inspired by the work presented in \cite{zhang2019self}, where an attention mechanism is used to learn image features, this work proposes a similar approach to learning point cloud features but, instead of only one layer, this work proposes a network of stacked attention layers. 

The proposed attention network (AN) comprises four attention layers (AL). The network was limited to four due to computational constraints. Each AL represents a dot-product attention mechanism as shown in Fig.\ref{fig:attention_network}, which comparatively with additive attention \cite{bahdanau2014neural} is much faster and more space-efficient due to being  computationally  implementable through highly optimized matrix multiplication code \cite{NIPS2017_3f5ee243}. 

The AL has an input ($X$) with shape [${c\,\times\,h \,\times\, \delta }$] and an output ($Y$) with the same shape.  The input is mapped into new features spaces $K = W_Kx$, $Q = W_Qx$ and $V = W_Vx$  with $K$, $Q$ and $V$ with shape [${c \times n}$],  where $ n= h \times \delta $. The matrices $W_K$, $W_Q$ and $W_V$ are learned during training using convolutional layers.   Furthermore,  $K$ and $Q$  are used to compute the attention map $A_m \in \mathbb{R}^{c \times c} $ (\ref{eq:attention_map}), which  improves the feature representation capabilities \cite{feng2020point}:

\begin{equation}
     {\renewcommand{\arraystretch}{1.5}% for the vertical padding
     A_m = \frac{e^{Q^TK}}{ \sum_{n=1}^{n} e^{Q^TK}.} 
     }
    \label{eq:attention_map}
\end{equation}
Finally, the output is computed using  (\ref{eq:attention_output}),  where $\gamma$ is a trainable scalar,
 \begin{equation}
     {\renewcommand{\arraystretch}{1.5}% for the vertical padding
     y = x + \gamma A_m V.
     }
    \label{eq:attention_output}
\end{equation}

\subsection{Output Layer}

The output comprises three main operations. Firstly, max pooling is applied to the input, converting the tensor from shape $[c \times h \times \delta ] $ to $[h \times \delta]$. Secondly, the tensor is flatten assuming the final shape $[1 \times m]$, where $m = h \times \delta$. Finally, the flattened tensor is normalized using layer normalization. 

\begin{figure*}[t]
\includegraphics[width=1\textwidth, trim={0cm 0cm 0cm 0cm},clip]{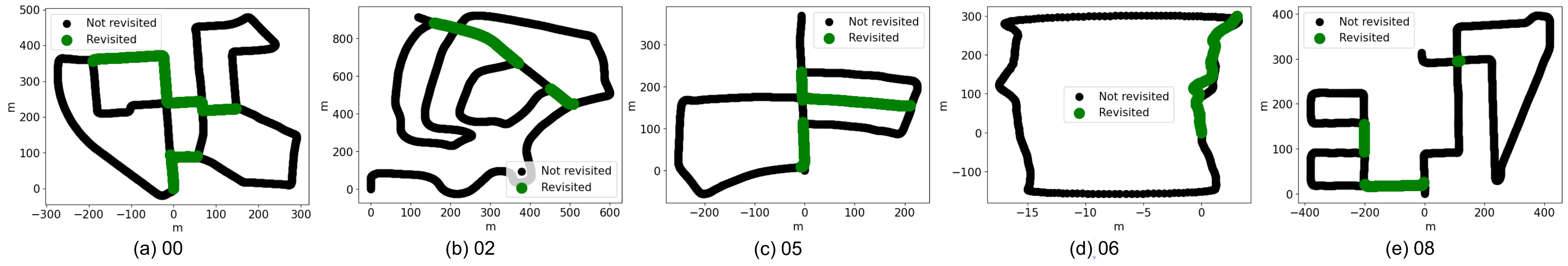}
\caption{KITTI Odometry Benchmark sequences used in this work, with the loop segments (in green) and segments with no loops (in black) highlighted: a) \textit{00}, b) \textit{02}, c) \textit{05}, d) \textit{06} and e) \textit{08}. Sequences \textit{00}, \textit{02}, \textit{05} and \textit{06} contain segments that are always revisited  from the same direction, while sequence 08 contains segments that are revisited from opposing direction. For example, in (a) black segments represent  locations that were not revisited, while green segments represent locations that were revisited.}
\label{fig:kitti}
\end{figure*}

%%%========= New Section =========
\section{EXPERIMENTAL EVALUATION} \label{sec:experiments}

The proposed approach was evaluated on the  KITTI  odometry dataset, using only the sequences with revisited segments. The ground-truth data were generated on purpose for this work, using a cross-validation scheme to obtain robust results. Additionally, an ablation study was conducted to assess the performance of the stacked attention layers and their relevance in a place recognition setting. And finally, the proposed approach was compared with PointNetVlad, a state-of-the-art place recognition approach. 

\subsection{KITTI Odometry Benchmark}

The KITTI Odometry Benchmark \cite{Geiger2012CVPR} is a collection of 22 sequences, containing point clouds, images, and GPS recordings of inner-city traffic, residential areas, highway scenes, and countryside roads around Karlsruhe, Germany. The sequences are split in training set  (sequences \textit{00} to \textit{10})  and test set (sequences \textit{11} to \textit{21}).  
The point clouds are recorded by a Velodyne 64 HDL sensor at 10 Hz, placed on the center of the car’s roof. Ground truth poses are provided by an RTK GPS sensor. 

However, not all sequences contain segments that are revisited. Only the sequences \textit{00}, \textit{02}, \textit{05}, \textit{06}, and \textit{08} have substantial segments with revisited paths, which were used in this work to train and validate the proposed approach. 
Figure \ref{fig:kitti} outlines the sequences that were used and highlights with green the revisited segments. 
Moreover, sequences \textit{00}, \textit{02}, \textit{05}, \textit{06} have segments that are revisited from the same direction; while sequence 08 has segments that are revisited from the opposite direction, which is particularly useful to validate viewpoint invariance. 

%%%------------- New Sub-section
\subsection{Ground-truth Data and Evaluation}
\label{sec:gte}

The ground truth data was generated based on the following: the selected sequences were split into a set of query frames $\{F_q\}$ and a set of reference frames $\{F_r\}$. Reference frames $F_r$ are point clouds from places that were not yet visited, while query frames $F_q$ are point clouds from places that were already visited (\ie, revisited segments).  
Thus, to train AttDLNet,  $F_q$ and $F_r$ are considered a positive pair when a loop exists and a negative pair otherwise. The loops $L=\{l | l \in [0,1]\}$ are defined as follows:

\[ l(P_q,P_{r}) =
  \begin{cases}
    1  &, \quad \|P_q - P_{r}\| < r_{th}\\
    0  &, \quad \text{otherwise}
  \end{cases}
\]

\noindent where $P_q$ and  $P_{r}$ are the position in the physical world of $Fq$ and $Fr$, respectively, with  $P_q, P_{r} \in \mathbb{R}^3$. And $r_{th}$ represents the boundary (in meters) within which the loop is considered valid.

For this work $r_{th}$ is set to 6\,m: \ie, a loop exists whenever $F_q$ and $F_r$ are within a range of 6\,m. Table \ref{tab:sequence} contains the frames length and the number of loops detected in each sequence. The ground truth loops are illustrated in Fig. \ref{fig:kitti}.

To obtain conclusive and robust results, the proposed approach was evaluated using precision (Eq. \ref{eq:precision}), recall (Eq. \ref{eq:recall}), and the $F_1$\,score (Eq. \ref{eq:f1}) in a 5-fold cross-validation scheme, where each fold represents a different sequence. In the context of this work, precision measures the fraction of the frames that were correctly retrieved among all retrieved frames, while recall measures the fraction of the frames that were correctly retrieved within all loops.

 \begin{equation}
     {\renewcommand{\arraystretch}{1.5}% for the vertical padding
     \text{precision} = \frac{\text{TP (\# Retrieved Loops)}}{\text{TP + FP (\# Retrieved Frames)}}
     }
    \label{eq:precision}
\end{equation}

 \begin{equation}
     {\renewcommand{\arraystretch}{1.5}% for the vertical padding
     \text{recall} = \frac{\text{TP (\# Retrieved Loops)}}{\text{TP + FN (\# Loops)}}
     }
    \label{eq:recall}
\end{equation}

 \begin{equation}
     {\renewcommand{\arraystretch}{1.5}% for the vertical padding
     F_{1} \, \text{score} = 2 \times \frac{\text{precision} \times \text{recall}}{\text{precision + recall}}
     }
    \label{eq:f1}
\end{equation}

\begin{table}[t]
  \centering
  \caption{\CP{Dataset and Loop information} Number of frames and loops in the dataset.}
  {\renewcommand{\arraystretch}{1.5}% for the vertical padding
	\begin{adjustbox}{max width=\columnwidth}
        \begin{tabular}{p{0.30\columnwidth}p{0.10\columnwidth}p{0.10\columnwidth}p{0.10\columnwidth}p{0.10\columnwidth}p{0.10\columnwidth}}
        \toprule
        Sequence: & 00     & 02     & 05     & 06     & 08 \\
        \midrule
		\midrule
         Length [frames]: &  4051 &  4661 & 2761 &   1101  & 4071 \\
         Loops [frames]:  &  801 &  306  & 489  &   265   &  329 \\ 
        \bottomrule
        \end{tabular}%
    \end{adjustbox}
    }
  \label{tab:sequence}%
\end{table}%

\begin{figure*}[t]
\includegraphics[width=1\textwidth, trim={0.3cm 0cm 1.2cm 0cm},clip]{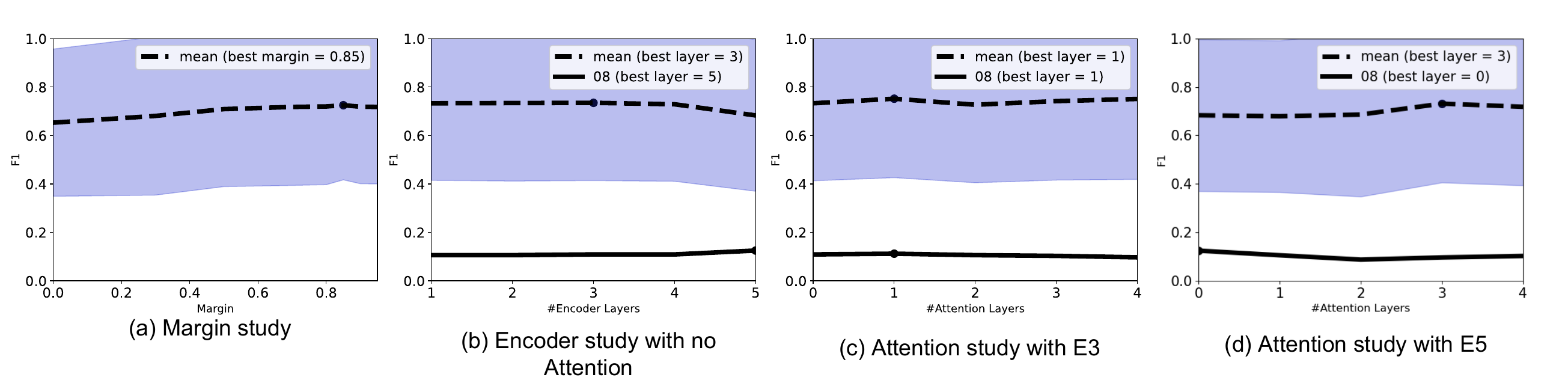}
\caption{AttDLNet ablation studies which include (a) margin, (b) encoder with no attention, (c) attention with E3 encoder configuration, and (d) attention with E5 encoder configuration.}
\label{fig:studies}
\end{figure*}

\subsection{Implementation and Training Setup}
\label{sec:its}

The proposed approach was implemented based on the PyTorch \cite{NEURIPS2019_bdbca288} frameworks and validated on a hardware setup with an NVIDIA GFORCE GTX1070Ti GPU and an AMD Ryzen 5 CPU with 32 GB of RAM. 
The network was trained using the Adam optimizer \cite{kingma2014adam} with a learning rate of 0.001, while the loss was computed based on a cosine similarity function, which measures the similarity between $F_q$ and $F_r$ of a positive pair and a negative pair. The loss function was implemented as follows:

\begin{equation}
    \mathcal{L}(F_q,F_r) =
  \begin{cases}
    1-cos(F_q,F_r), & \text{Positive pair} \\
    max(0,cos(F_q,F_r) - \epsilon), & \text{Negative pair}\\
  \end{cases}
\end{equation}

\noindent where $\epsilon$ represents a margin parameter, which was obtained through a thorough assessment (see Fig. \ref{fig:studies}.a). From this assessment, the margin value that returned the highest performance was 0.85, which was obtained by computing the mean F1 value of AttDLNet's retrieval performance on the sequences \textit{00}, \textit{02}, \textit{05}, \textit{06}, and \textit{08}.    

During the training phase of the proposed approach, both positive and negative pairs were randomly selected, however, to enforce the dissimilarity between frames, the negative pairs were formed by frames that are at least 20\,m apart (\ie, $r_{th} > 20$\,m) (as illustrated in Fig. \ref{fig:triplet}).

\begin{table}[t]
  \centering
  \caption{F1 scores of the AttDLNet's encoder and attention ablation study. The attention study uses the encoder architecture that has the highest mean F1, which are E3  and E5.}
  {\renewcommand{\arraystretch}{1.5}% for the vertical padding
	\begin{adjustbox}{max width=\columnwidth}
        \begin{tabular}{p{0.13\columnwidth}p{0.13\columnwidth}p{0.13\columnwidth}p{0.13\columnwidth}p{0.13\columnwidth}p{0.13\columnwidth}p{0.13\columnwidth}p{0.13\columnwidth}}
        \toprule
        Seq.   & 00        & 02    & 05    & 06    & 08    & mean & FPS\\
        \midrule
		\midrule
         E1        &  0.90    & 0.82 & 0.86 & 0.98   & 0.11 &   0.73  & 368\\
         E2        &  0.93    & 0.78 &  0.86 & \textbf{1.00} & 0.11 & 0.73 & 260\\
         E3        &  0.92    & 0.78  & \textbf{0.87} & \textbf{1.00 }& 0.11 & \textbf{0.74} & 103 \\
         E4        &  0.\textbf{94}    & \textbf{0.79} &  0.82 & 0.98   &0.11 & 0.73 & 52\\
         E5        &  0.94    & 0.55 & 0.85 &  0.95   &\textbf{0.13} & 0.68  &  42\\
         \midrule
         E3A0        &  0.92    & 0.77 & 0.87 & \textbf{1.00}   & \textbf{0.11} &   0.73  & 103\\
         E3A1        &  \textbf{0.95}    &  \textbf{0.82} &  0.88 & \textbf{1.00} & \textbf{0.11} & \textbf{0.75} & 78\\
         E3A2        &  0.94    & 0.74 & 0.87 & 0.99 & \textbf{0.11} & 0.72 & 63 \\
         E3A3        &  0.94    & 0.80 &  0.88 & 0.99   &0.10 & 0.74 & 53\\
         E3A4        &  0.94    &  \textbf{0.82} & \textbf{0.90} &  \textbf{1.00}   &  0.10 & \textbf{0.75}  &  50\\
          \midrule
         E5A0        &  0.94    & 0.55 & 0.85 & 0.95   & \textbf{0.13} &   0.68 & 42\\
         E5A1        &  0.91    & 0.62 & 0.77 & \textbf{0.99}   &0.11 & 0.68 & 78\\
         E5A2        &  0.94    & 0.54 & \textbf{0.89} & 0.98   &0.09 & 0.69 & 63 \\
         E5A3        &  0.94    & \textbf{0.79} &  0.85 & 0.98  &0.10 &  \textbf{0.73} & 53\\
         E5A4        &  \textbf{0.95}    & 0.69 & 0.86 &  \textbf{0.99}  &  0.10 & 0.72  &  50\\
         \midrule
         
        \bottomrule
        \end{tabular}%
    \end{adjustbox}
    }
  \label{tab:ablation}%
\end{table}%

\begin{figure*}[t]
\includegraphics[width=1\textwidth, trim={0cm 0cm 0cm 0cm},clip]{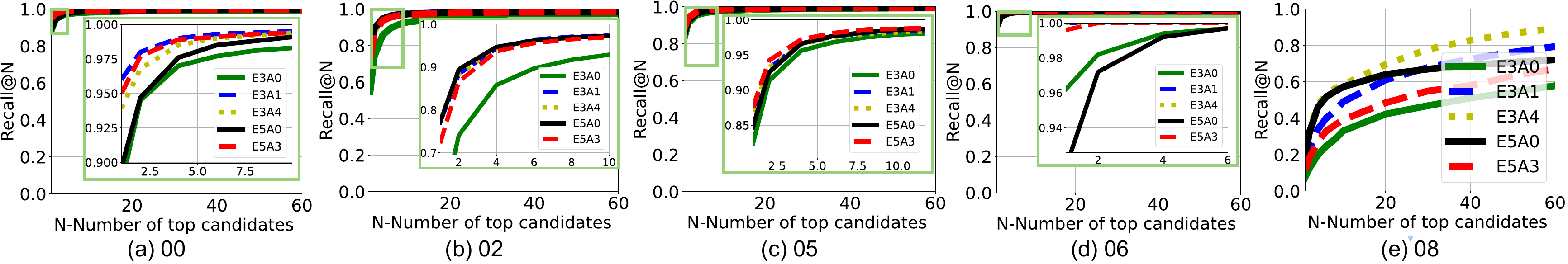}
\caption{Recall scores of E3A0, E3A1, E3A4, E5A0 and E5A3 AttDLNet's architectures on sequence: (a) \textit{00}, (b) \textit{02}, (c) \textit{05}, (d) \textit{06} and (e) \textit{08}.}
\label{fig:recall}
\end{figure*}

\subsection{Ablation Study} 
\label{sec:ablation}

The ablation study aimed to study the performance of stacked attention layers in place recognition. 
The study is divided into two phases: first, the study focuses on the encoder network with no attention in the proposed network to establish a baseline; the second phase uses this baseline architecture to study the attention network. Specifically, the study consists in exploiting the contribution of each attention layer to the overall performance. 

For the purpose of this study AttDLNet's architecture is identified by $E_{x}A_{y}$, where $E_x$ represents the encoder configuration with $x \in [1,2,3,4,5]$, and $A_y$ the attention network configuration with $y \in [0,1,2,3,4]$.  For example, $E_{3}A_{1}$ is a network that has three encoder layers and one attention layer (see Fig. \ref{fig:overview_pipeline}). Another example is E3A0, which represents the aforementioned baseline, where the encoder has three layers and no attention layer is present in the network. 

The results of these studies are presented in Table \ref{tab:ablation}, where only the first nearest neighbors predictions were considered. The mean values are presented in Fig. \ref{fig:studies} (b), (c) and (d). The results from the encoder study indicate that the middle encoder layers have in general higher performance, being $E_3$ the configuration with the best mean performance, while $E_5$ is the best configuration on sequence \textit{08}. The first choice for the baseline should be $E_3$, however, since $E_5$ has the highest score in sequence \textit{08}, which is an extremely challenging scenario since all revisits are from the opposing direction, both  $E_3$ and $E_5$ were selected as baseline configurations.

Regarding the performance of the $E_3$ baseline, the best performance is obtained using one attention layer(\ie, $A1$) and four stacked attention layers ($A4$). While the performance of the $E_5$ baseline, the best attention configuration is obtained using three stacked layers $A_3$. Moreover, the best performance on sequence \textit{08} is obtained by $E_{5}A_{0}$ which means that, in terms of rotation invariance, deeper networks are more robust.

All three aforementioned AttDLNet configurations (\ie, $E_{3}A_{1}$, $E_{3}A_{4}$, and $E_{5}A_{3}$), as well as the configurations without attention ($E_{3}A_{0}$ and $E_{5}A_{0}$), are further compared and studied in terms of retrieval performance. 

\subsection{Retrieval Performance} 

The proposed approach was further evaluated on a retrieval task. The evaluation is conducted on the configurations $E_{3}A_{1}$, $E_{3}A_{4}$, $E_{5}A_{3}$, $E_{3}A_{0}$ and $E_{5}A_{0}$, considering as acceptable a range of $N \in [1,2,4,6,8,10,20,30,40,50,60]$ top candidates. It is expected that as the number of top candidates grows, more true loops are retrieved. To measure this, we used recall (Eq. \ref{eq:recall}). Furthermore, we compare the configuration $E_{3}A_{1}$ and $E_{3}A_{4}$ of AttDLNet with PointNetVlad \cite{angelina2018pointnetvlad} using 80k points as input and an output descriptor with a size of 1024 elements, which is the same size as AttDLNet.
% %is progressively increased in order to evaluate  AttDLNet's capacity to map frames originate from the same physical space close in the feature space. 

%To perform this study, the network architectures that have the highest performance in the ablation study were selected, which correspond to the configurations  $E3A0$, $E3A1$, $E3A4$, $E5A0$, and $E5A3$; while the $N$ top candidate set include $N \in [1,2,4,6,8,10,20,30,40,50,60]$.

The results of these studies are shown in Fig.\ref{fig:recall}, indicating that network configurations with attention have in general higher performance. When analyzing results from sequence \textit{08},  (where the ablation studies show that $E_{5}A_{0}$ has the highest performance) as $N$ grows, networks with attention outperform networks with no attention. 

The results of the comparison are presented in table \ref{tab:comaprison}, which indicates that our approach outperforms PointNetVlad using the same output dimension. Furthermore, AttDLNet achieves higher performance with higher efficiency, when comparing the frame rates of both approaches: AttDLNet has 78 and 50 FPS, and PointNetVlad has 9.5 FPS.   

%This demonstrates that despite ablation studies indicating (for N=1) that deeper networks are more ration invariant, when N grows, networks with attention have a steeper performance grows, outperforming with N greater than 20 networks with no attention.  Thus, the results indicate that attention improves place recognition in general and in particular when the number of retrieval candidates grows. 

\begin{table}[t]
  \centering
  \caption{F1 score results of PointNetVlad(PNV)(number of points) }
  {\renewcommand{\arraystretch}{1.5}% for the vertical padding
        \begin{tabular}{p{0.2\columnwidth}ccccccc}
        \toprule
        Seq.   & 00        & 02    & 05    & 06    & 08    & mean & FPS\\
        \midrule
		\midrule
         AttDLNet($E_{3}A_{1}$)        & 0.95    & 0.82 &  0.88 & 1.00      & 0.11 &    0.75    &   78\\
         AttDLNet($E_{3}A_{4}$)        & 0.94    & 0.82 &  0.90 & 1.00      & 0.10 &    0.75    &   50\\
         %PNV(10k)     &  0.24    & 0.19 &  0.41 & 0.27      & 0.19 &    0.27    &   67\\
         %PNV$_{254}$(10k)     &  0.24    & 0.19 &  0.41 & 0.27      & 0.19 &    0.27    &   67\\
         %PNV(80k)     &  0.26    & 0.18  & 0.43 &0.25       & 0.26 &    0.28    &   9.7 \\
         PNV$_{1024}$(80k)     &  0.88    & 0.57  & 0.82 & 0.64       & 0.36 &    0.65    &   9.7 \\
        \bottomrule
        \end{tabular}%
    }
  \label{tab:comaprison}%
\end{table}%

%%%========= New Section =========
\section{CONCLUSIONS}
In this work, a retrieval-based place recognition approach is proposed, which includes the novel AttDLNet network. AttDLNet is a point cloud-based DL network, which resorts to an encoder and an attention network to improve descriptiveness. The network converts point clouds to range-based proxy representation and concurrently exploits an attention network to selectively focus on the most relevant features.
 
The experimental evaluation (carried out on the KITTI dataset) shows that middle encoder layers have higher mean performance than early or deeper layers. However, in the particular case when places are revisited from the opposite direction, deeper layers have higher performance, i.e., more orientation invariant. When considering the place recognition pipeline, results show that the use of the AttDLNet network allows achieving high place recognition performance, being able to outperform a well-established approach.  
 
%%%========= New Section =========
\section*{ACKNOWLEDGMENTS}
This work has been supported by the projects MATIS-CENTRO-01-0145-FEDER-000014 and SafeForest CENTRO-01-0247-FEDER-045931, Portugal. It was also partially supported by FCT through grant UIDB/00048/2020 and under the PhD grants with reference SFRH/BD/148779/2019 and 2021.06492.BD.

%%%%%%%%%%%%%%%%%%%%%%%%%%%%%%%%%%%%%%%%%%%%%%%%%%%%%%%%%%%%%%%%%%%%%%%%%%%%%%%%

\bibliographystyle{IEEEtran}

\bibliography{ref}

\end{document}